\documentclass[lettersize,journal]{IEEEtran}
\usepackage{amsmath, amssymb, booktabs, makecell}
\usepackage{algorithmic}
\usepackage{algorithm}
\usepackage{array}
\usepackage[caption=false,font=normalsize,labelfont=sf,textfont=sf]{subfig}
\usepackage{textcomp}
\usepackage{stfloats}
\usepackage{url}
\usepackage{verbatim}
\usepackage{graphicx}
\usepackage{cite}
\usepackage[colorlinks=true, linkcolor=blue, citecolor=blue, urlcolor=blue]{hyperref}
\usepackage{color}
\usepackage{moreverb,url}
\usepackage{tabularx}
\usepackage{caption}
\usepackage{xcolor}
\usepackage{float}
\usepackage[normalem]{ulem}
\usepackage{multicol}
\begin{document}

\title{Large Language Model Enhanced Differentiable Trajectory Planning for IoT-Enabled Autonomous Driving}

\author{Shihao Zhang, Jing Yang, Ziyu Song, Zheng Lin, Sunil Prajapat, Zhaochen Xia, Hemant Ghayvat, Haitao Ding, Lip Yee Por, Ashok Kumar Das
\thanks{This work was supported by the China Automobile Industry Innovation and Development Joint Fund under Grant No. U1864206, and China Shenzhen Major Science and Technology Projects under Grant No. ZDCY20250901101003004. ({Corresponding author: Haitao Ding}.)}

\thanks{Shihao Zhang, Ziyu Song, Zhaochen Xia and Haitao Ding are with the State Key Laboratory of Automotive Simulation and Control, Jilin University, Changchun 130000, China (e-mail: shihao24@mails.jlu.edu.cn; songziyu@jlu.edu.cn; xiazc24@mails.jlu.edu.cn; dinght@jlu.edu.cn).}

\thanks{Jing Yang and Por Lip Yee are with the Center of Research for Cyber Security and Network (CSNET), Faculty of Computer Science and Information Technology, Universiti Malaya, 50603 Kuala Lumpur, Malaysia (e-mail: s2147529@siswa.um.edu.my; porlip@um.edu.my).}

\thanks{Zheng Lin is with the Interdisciplinary Centre for Security, Reliability and Trust (SnT), University of Luxembourg, Luxembourg (e-mail: zhenglin@ieee.org).}

\thanks{Sunil Prajapat and Hemant Ghayvat are with the IMT, Department of Humanities and Technology, Roskilde University, Roskilde, Denmark (e-mail: sunilprajapat645@gmail.com; hghayvat@ruc.dk).}

\thanks{Ashok Kumar Das is with the Center for Security, Theory and Algorithmic Research, International Institute of Information Technology, Hyderabad 500 032, India, and also with the Department of Computer Science
and Engineering, College of Informatics, Korea University, 145 Anam-ro, Seongbuk-gu, Seoul 02841, South Korea (e-mail: iitkgp.akdas@gmail.com, ashok.das@iiit.ac.in).}
}

\markboth{IEEE Internet of Things Journal}%
{Shell \MakeLowercase{\textit{et al.}}: A Sample Article Using IEEEtran.cls for IEEE Journals}

\maketitle

\begin{abstract}
Autonomous driving planning is a key component of IoT-enabled intelligent transportation systems, requiring vehicles to generate safe, efficient, and executable trajectories in complex urban environments from multi-source contextual information. While imitation learning (IL) has shown promise on large-scale datasets, IL-based planners still suffer from limited coverage of complex long-tail interactions, weak consistency with downstream constrained refinement, and insufficient use of high level scene semantics under real time constraints. To address these issues, this paper proposes a large language model (LLM) enhanced differentiable trajectory planning framework for IoT-enabled autonomous driving. Specifically, we introduce a surrounding agent centric data augmentation strategy to reorganize surrounding agent trajectories as additional planning supervision, thereby improving the training distribution without collecting additional raw data. We further design a complexity-aware asynchronous LLM-based semantic enhancement module to extract scene-related high-level semantic features with controlled online overhead. In addition, a differentiable optimization module is incorporated to refine generated trajectories with explicit residual penalties while backpropagating optimization gradients to the upstream planner. 
Experiments show that the proposed method achieves the best overall scores of 83.63 and 78.29 on the nuPlan closed-loop nonreactive and reactive Hard20 benchmarks, respectively, and CARLA-ROS tests further verify its online deployment and real time closed-loop execution capability.
\end{abstract}

\section{Introduction}

\IEEEPARstart{A}{utonomous} driving systems are a key component of IoT-enabled intelligent transportation services~\cite{gong2024cooperative,yang2026explainable,lin2026gapsl,fang2024ic3m}, where vehicles integrate multi-source contextual information, including surrounding traffic participants, maps, traffic states, and navigation instructions, to generate safe, efficient, and socially compliant trajectories in complex environments~\cite{song2024enhanced,song2018safety,zhang2026reliable}. Traditional rule-based planners rely on manually designed rules and objective functions, which often limits their adaptability and generalization in dynamic and uncertain traffic scenarios. To address these limitations, researchers have explored model-based approaches, including imitation learning (IL)~\cite{song2025smart}, reinforcement learning (RL)~\cite{yang2026qugrid,duan2025leed,zhang2025robust}, and hybrid methods~\cite{lee2025episodic}, improving the flexibility of autonomous driving systems in complex environments.

Among these approaches, imitation learning has attracted considerable attention because it can leverage expert demonstrations to learn complex driving behaviors, such as lane changing, obstacle avoidance, and interactions with surrounding traffic participants. Several IL-based planning models~\cite{renz2022plant,scheel2022urban,cheng2024rethinking} have achieved promising performance on large scale real world datasets such as nuPlan~\cite{caesar2021nuPlan}. However, existing IL-based methods still face several critical challenges in practical deployment for complex urban mobility services \cite{yang2025neuroagent}.

\begin{figure}[t]
    \centering
    \includegraphics[width=0.5\textwidth]{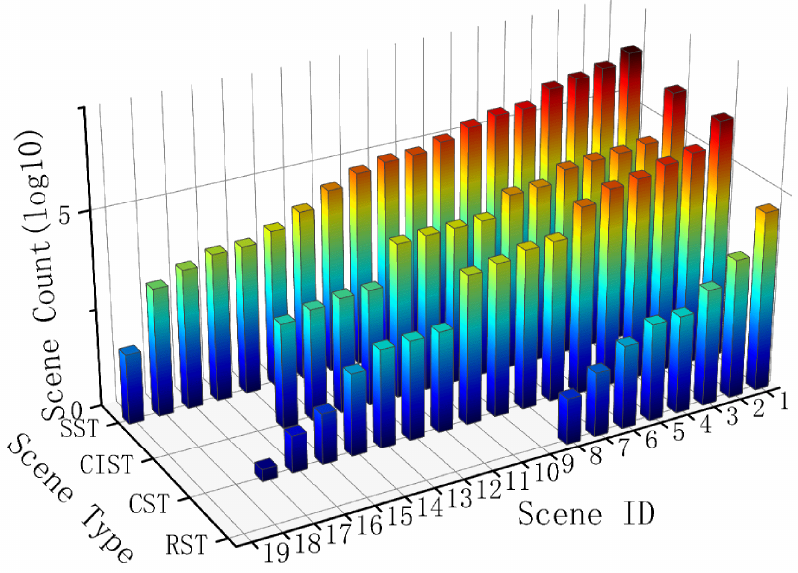}
    \caption{Scenario distribution statistics of the nuPlan dataset. 
    The dataset shows a typical long-tail distribution: Simple scenarios dominate,
    while constrained, complex, and especially rare scenarios are less frequent 
    but critical for evaluating robustness and generalization. SST: Simple Scenario Types; CIST: Constrained Interaction Scenario Types; CST: Complex Scenario Types; RST: Rare Scenario Types. }
    \label{fig:scenario_statics}
\end{figure}

First, imitation learning relies heavily on expert driving data during training. Although high quality public datasets such as nuPlan and the Waymo Open Dataset~\cite{sun2020scalability} are available, highly interactive, complex, and rare scenarios remain relatively scarce. From a data centric IoT intelligent transportation perspective, this long-tail imbalance may limit the robustness and generalization of learning-based planners in safety critical urban mobility services. Our statistical analysis of approximately $2 \times 10^7$ nuPlan scenarios further confirms this issue (in Fig.~\ref{fig:scenario_statics}): simple scenarios dominate the dataset, whereas complex and rare scenarios account for only a small proportion. Since prior studies have also suggested that data quality can be more critical than data volume for model performance~\cite{bronstein2023embedding}, enriching complex traffic behavior samples becomes important for improving the practical utility of autonomous driving planning models.

Second, in highly interactive traffic environments, pure imitation learning models often struggle to maintain stable and robust performance, and their generated trajectories still leave room for improvement in safety, feasibility, and comfort~\cite{huang2022multi,gao2021interacting,mo2022multi}. To improve controllability, existing methods commonly combine IL planners with rule-based post processing, filtering, or hybrid planning strategies~\cite{dauner2023parting,cheng2024pluto}. However, these mechanisms are usually applied at the output stage and cannot sufficiently feed safety boundaries, kinematic constraints, or interaction risks back into the upstream planner during training. This may lead to inconsistency between generated candidate trajectories and downstream constrained refinement objectives. In addition, many existing methods loosely couple prediction and planning, making it difficult for prediction results to fully support planning decisions and for planning objectives to guide trajectory generation. Although unified prediction and planning frameworks can implicitly model interactions, they may still lack sufficient stability and robustness in safety critical scenarios~\cite{huang2023differentiable}.

Furthermore, imitation learning models are prone to performance degradation in out of distribution long-tail scenarios due to limited generalization capability. Large language models (LLMs)~\cite{lin2025pushing,fang2024automated,lin2026hsplitlora,duan2025llm,fang2026hfedmoe} have shown promising commonsense reasoning and zero-shot generalization ability, offering a new opportunity for enhancing scene understanding in complex driving scenarios. However, directly applying LLMs to autonomous driving planning remains challenging. Critical traffic information, such as road topology, dynamic interactions, and fine-grained spatiotemporal context, is difficult to represent fully in compact text. Moreover, large parameter scale and inference overhead of LLMs further make direct high frequency deployment difficult for real time closed-loop planning in IoT-enabled intelligent transportation systems \cite{lin2025pushing}.

To address the above challenges, we develop a unified IL-based planning framework by integrating surrounding agent centric data augmentation, complexity-aware asynchronous LLM-based semantic enhancement, and residual-based differentiable optimization. The main contributions of this work are summarized as follows:
\begin{enumerate}
    \item We propose an LLM enhanced IL-based planning framework with differentiable optimization. The proposed framework leverages a complexity-aware asynchronous LLM to provide high level scene semantic guidance, while a differentiable optimization module imposes explicit constraints on the output trajectories and backpropagates optimization gradients to the upstream planning model, thereby improving scene understanding and closed-loop planning performance in complex scenarios.
    \item We develop a surrounding agent centric data augmentation strategy. By reorganizing the real trajectories of surrounding traffic participants in complex scenarios as additional planning supervision, the proposed strategy increases the proportion of high value training samples, which improves the training distribution and enhances the model’s generalization capability in complex scenarios without collecting additional raw data.
    \item Comprehensive experiments on the nuPlan closed-loop benchmarks demonstrate that the proposed method achieves the best overall scores of 83.63 and 78.29 on the nuPlan nonreactive and reactive challenges on the Hard20 split, respectively. In addition, experiments on the real time CARLA-ROS software-in-the-loop (SIL) platform further verify the online deployment feasibility and real time closed-loop execution capability of the proposed framework.
\end{enumerate}

\section{RELATED WORK}\label{sec2}

\subsection{Data Augmentation for Autonomous Driving}

Existing data augmentation methods for autonomous driving can be broadly categorized into perception level, scene level, and trajectory level approaches. Perception level methods increase input diversity through image transformations, noise perturbations, or weather variations~\cite{yang2025survey}. Scene level methods enrich traffic scenarios through simulation reconstruction, digital twins, or generative models, including LLM-based scenario generation~\cite{li2024chatgpt}. Trajectory level methods reuse expert trajectories, behavioral patterns, or interaction processes as planning-related demonstrations~\cite{mirkhani2024augmenting}.

Despite these advances, non-ego motion records contained in real driving logs remain underexploited as planning demonstrations. Our surrounding agent centric data augmentation strategy therefore reindexes selected surrounding vehicles as planning subjects, allowing one traffic episode to provide multiple agent centric training samples without additional data collection.

\subsection{IL-based Planning}

Existing IL-based planners mainly fall into three categories. The first category consists of pure IL-based planners, which directly map scene representations to planned trajectories such as PlanT and PlanTF \cite{renz2022plant,cheng2024rethinking}. The second category focuses on interaction-aware or joint prediction and planning methods, which explicitly model the coupling between ego decisions and surrounding agent behaviors to handle complex dynamic scenarios, as exemplified by GameFormer~\cite{huang2023gameformer}. 
The third category~\cite{dauner2023parting,cheng2024pluto} combines IL with output-stage trajectory enhancement. In this line of work, post processing evaluates or ranks candidate trajectories after inference without modifying the trajectory generation process, whereas trajectory refinement directly adjusts the geometry or state sequence of a selected trajectory to improve performance of the trajectory.

Inspired by recent progress in IL-based planning~\cite{huang2023differentiable,cheng2024pluto,amos2017optnet,amos2018differentiable}, we employ a residual-based differentiable optimizer to refine the selected initial ego trajectory using the predicted trajectories of surrounding agents as conditional inputs. Since the optimization process is differentiable, gradients associated with the residual objectives can be backpropagated through the solver to the upstream planning network, allowing the optimization objectives to guide initial trajectory generation during training.

\subsection{LLM for Autonomous Driving}

Existing studies can be broadly grouped into three directions. 
First, some works combine natural language instructions with multimodal perception inputs for closed-loop or goal-oriented driving. 
For example, LMDrive integrates multimodal sensor inputs and language instructions for closed-loop end-to-end driving, LeGo-Drive uses language guided goal representations for closed-loop trajectory generation, and DriveLM formulates driving scene understanding as graph-based visual question answering for perception, prediction, and planning reasoning~\cite{shao2024lmdrive,paul2024lego,sima2024drivelm}. 
Second, another line focuses on rule understanding, decision reasoning, and interpretable driving cognition. 
Driving with Regulation incorporates traffic laws, regulations, and safety requirements into driving decision making through retrieval augmentation and language reasoning, while Reason2Drive provides chain-based reasoning data for interpretable autonomous driving~\cite{nie2024reason2drive}. 
Third, survey studies such as LLM4AD summarize representative LLM-based paradigms and show their potential in scene understanding, rule reasoning, and decision support~\cite{cui2026llm4ad}. 
These studies indicate that LLMs are more suitable as sources of high level knowledge and semantic information, complementing downstream decision making and planning modules rather than replacing conventional planners.

Nevertheless, applying LLMs~\cite{lin2024splitlora,qu2025mobile,lin2025hierarchical,fang2024dynamic,lyu2023optimal} to autonomous driving remains challenging because high level language semantics do not directly match the continuous and geometry-constrained representations required for trajectory level planning, and closed-loop LLM inference still incurs considerable real time overhead. 
Recent LLM-assisted driving studies, including LMDrive~\cite{shao2024lmdrive} and AsyncDriver~\cite{chen2024asyncdriver}, motivate using language models as high level semantic feature providers for real time planning. 
Following this direction, this paper introduces an asynchronous LLM-based scene-associated feature extraction module for an IL-based real time planner. 
A lightweight scene complexity recognition mechanism further adaptively schedules semantic feature updates according to traffic density, interaction risk, navigation changes, and short term scene variations. 
The extracted semantic features are fused with scene encodings and coupled with differentiable trajectory refinement, enabling scene level semantic guidance to support trajectory proposal generation and optimization-aware refinement while keeping online computation under control.

\section{METHODOLOGY}\label{sec3}

\subsection{Problem Formulation}
This paper studies the decision making and motion planning problem for autonomous driving in urban traffic environments. Let the autonomous vehicle be denoted by $A_0$, and the other traffic participants by $\{A_i\}_{i=1}^{N_A}$. Let $s_i^t$ denote the state of the $i$-th traffic participant at time step $t$, and let $T_H$ and $T_F$ represent the historical observation horizon and the future planning horizon, respectively. At the current time step $t=0$, given the historical state sequence \(S=\{s_i^t\mid i=0,\dots,N_A,\;t=-T_H,\dots,0\}\), the map scene context \(M\), and the high level language information \(I\), the model is required to output a set of \(N_T\) multimodal initial planning trajectories for the ego vehicle, \(T_0=\{\tau_0^n\}_{n=1}^{N_T}\), together with their corresponding confidence scores, while simultaneously predicting the trajectories of surrounding traffic participants over the next $T_F$ time steps. Subsequently, the model further refines the ego trajectory using a differentiable optimizer.

\begin{figure*}[t]
    \centering
    \includegraphics[width=1.0\textwidth]{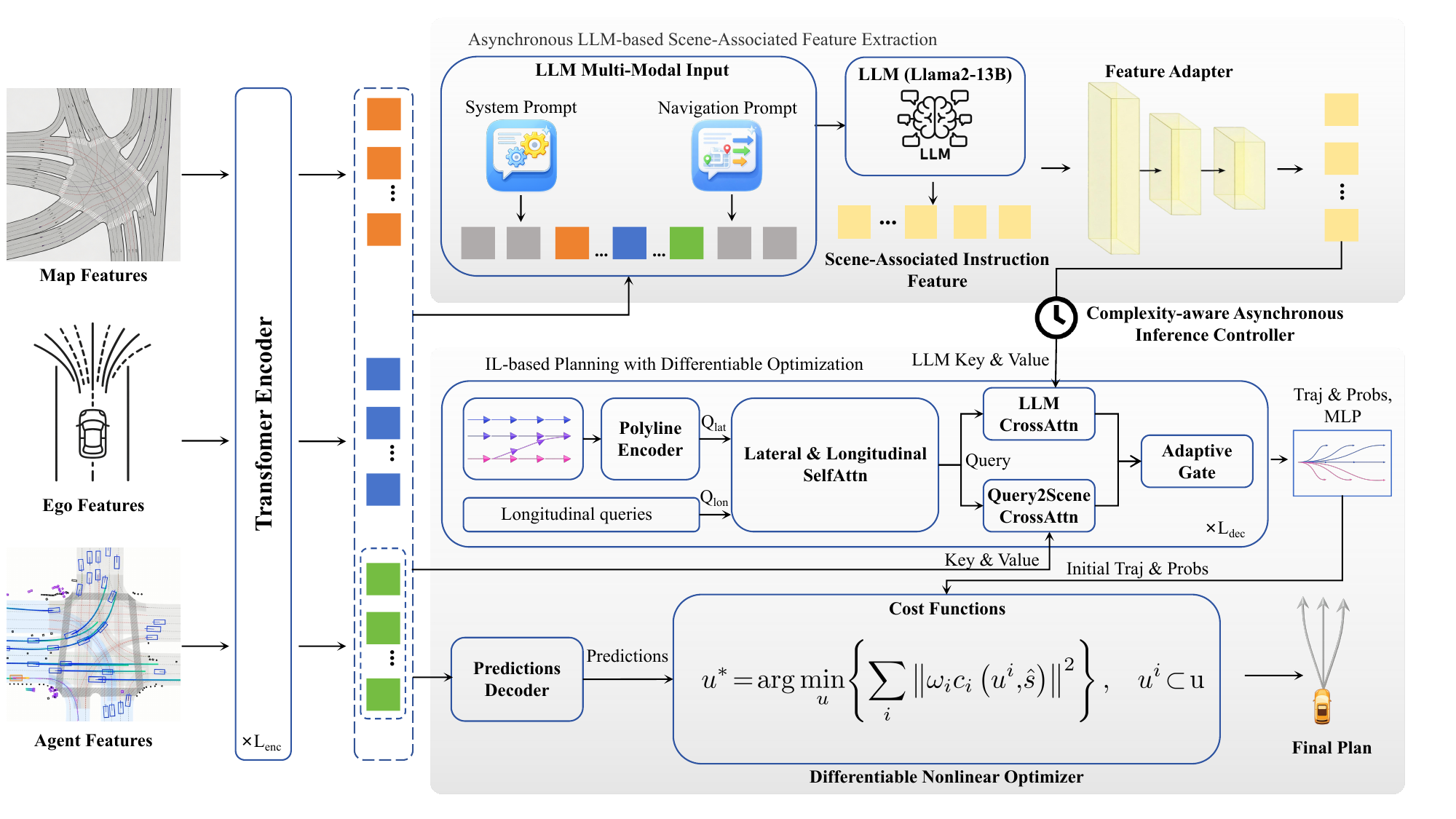}
    \caption{Overview of the proposed planning framework.}
    \label{fig:framework}
\end{figure*}

As illustrated in Fig.~\ref{fig:framework}, the proposed framework combines IL-based planning, asynchronous LLM-based semantic enhancement, and differentiable optimization. Map, ego, and surrounding agent features are first encoded by a Transformer encoder into a unified scene representation. The system prompt, navigation instructions, and encoded scene representation are then organized as multimodal LLM inputs to extract scene-associated instruction features, which are adapted and asynchronously updated before being injected into the planning module. The planning decoder generates multimodal initial trajectories and confidence scores, while the prediction decoder estimates future trajectories of surrounding agents. Finally, the differentiable nonlinear optimizer refines the highest confidence ego trajectory using predicted agent trajectories and residual-based cost terms. Since the optimizer is differentiable, planning-related gradients can be propagated back to the upstream planner during training, enabling joint learning of trajectory generation and optimization-aware refinement.

\subsection{Surrounding Agent Centric Data Augmentation}

IL-based planning typically uses only the ego vehicle's future trajectory as supervision, which limits the learned behavior prior to a single planning subject in each scene. To increase the coverage of complex interactions, we propose a surrounding agent centric data augmentation strategy that reuses the real trajectories of surrounding traffic participants as additional planning supervision.

The key idea is to reinterpret the same traffic scene from multiple agent centric perspectives while preserving its original semantics and interaction structure. Candidate surrounding vehicles are collected within a predefined screening radius and prioritized according to interaction relevant behaviors, including lane changing, intersection traversal or turning, low time to collision (TTC) interactions, high lateral acceleration, and high magnitude speed behaviors. To reduce noise from perception errors or incomplete tracking, we retain only target vehicles with valid observations, temporally continuous trajectories, and physically plausible motion patterns. After filtering, the scene is recentered around each retained target vehicle, so that a single traffic episode can provide multiple training samples with different planning subjects.

After selecting a target vehicle, the original scene is reformulated in a local coordinate system centered on that vehicle. Let \(p_c\) and \(\theta_c\) denote the current position and heading of the target vehicle, and let \(p\) denote the position of an arbitrary scene element in the global coordinate system. Its representation in the target-centered coordinate system is given by
\begin{equation}
\tilde{p}=R(-\theta_c)(p-p_c)
\end{equation}
where \(R(-\theta_c)\) denotes the two-dimensional rotation matrix determined by the target vehicle heading. Through this transformation, map elements, surrounding traffic participants, and the target vehicle state are consistently mapped into the new local frame, thereby reindexing the same scene as a new planning sample.

To maintain feature consistency with the downstream planner, missing dynamic states are approximated from trajectory sequences. Let \(\theta_t\) and \(v_t\) denote the heading angle and speed of the target vehicle at time step \(t\), respectively. The yaw rate and longitudinal acceleration are approximated as
\begin{equation}
\omega_t \approx \frac{\theta_{t+1}-\theta_{t-1}}{2\Delta t},
\quad
a_t \approx \frac{v_{t+1}-v_{t-1}}{2\Delta t}
\end{equation}
where \(\Delta t\) is the time interval between two consecutive time steps. In this way, surrounding agent trajectories are converted into additional planning samples that enrich complex behavior patterns without collecting new raw data.

\subsection{Asynchronous LLM-based Scene-Associated Feature Extraction}

To enhance the planner's ability to capture complex traffic scenes, navigation intent, and high level semantic constraints, we introduce an asynchronous LLM-based scene-associated feature extraction module for the downstream planning backbone, following the asynchronous semantic feature extraction paradigm~\cite{chen2024asyncdriver}. In this module, the LLM encodes the system prompt, navigation instructions, and structured scene representation into hidden semantic features. These features provide high level scene and instruction guidance to the planning decoder, while trajectory candidates, confidence scores, and final optimized trajectories are produced by the IL-based planner and the differentiable optimizer.

Since a general purpose pretrained large language model cannot be directly applied to autonomous driving planning, we adopt a lightweight adaptation strategy to align the language model with scene-semantic understanding and navigation instruction modeling, while introducing an asynchronous inference mechanism to reduce its online computational overhead.
Specifically, the high level language input \(I\) consists of a system prompt \(I_{\mathrm{sys}}\) and navigation instructions \(I_{\mathrm{nav}}\). The system prompt follows the fixed driving role template: ``Role: You are now an autonomous driving driver. I will provide you with the environment information, including ego vehicle information, surrounding agent information, and map information. Please extract scene-associated instruction features based on the given environmental information and routing instructions.'' The navigation instructions are generated from the reference path or route through a rule-based procedure and contain maneuver commands with distance information. The structured scene representation \(z_{\mathrm{scene}}\), extracted by the scene encoder from ego vehicle states, surrounding agent features, and map features, is formatted together with the tokenized system prompt and navigation instructions to form the multimodal LLM input
\begin{equation}
x_{\mathrm{llm}} = \Phi(I_{\mathrm{sys}}, I_{\mathrm{nav}}, z_{\mathrm{scene}})
\end{equation}
where \(z_{\mathrm{scene}}\) denotes the encoded scene representation and \(\Phi(\cdot)\) denotes the multimodal input construction process, including language tokenization, scene feature formatting, and the input alignment required by the LLM interface. In this way, the structured scene representation is converted into LLM-compatible contextual input, allowing the LLM to jointly encode navigation instructions and traffic scene information.

The large language model then encodes the multimodal input and outputs the hidden feature sequence
\begin{equation}
H_{\mathrm{llm}}=\{h_0,h_1,\ldots,h_{N_{\mathrm{tok}}-1}\}\in\mathbb{R}^{N_{\mathrm{tok}}\times D_{\mathrm{llm}}}
\end{equation}
where \(N_{\mathrm{tok}}\) denotes the number of valid input tokens after multimodal input construction and \(D_{\mathrm{llm}}\) denotes the hidden dimension of the LLM. Following the scene-associated instruction feature extraction strategy, the hidden feature of the last valid token is selected as the compact semantic representation:
\begin{equation}
f_{\mathrm{sem}} = h_{N_{\mathrm{tok}}-1}
\end{equation}

The feature adapter then maps \(f_{\mathrm{sem}}\) into the planner-compatible feature space:
\begin{equation}
F_{\mathrm{sem}} = \mathrm{Linear}(f_{\mathrm{sem}})
\end{equation}
where \(F_{\mathrm{sem}}\in\mathbb{R}^{D_{\mathrm{p}}}\) and \(D_{\mathrm{p}}=128\) in our implementation. The adapted feature captures navigation intent, scene context, and high level behavioral preferences associated with the current driving task, and is used as the scene-associated instruction feature for the downstream planning decoder.

Because LLM inference is computationally expensive, invoking it synchronously at every planning step would substantially increase online latency and reduce the practicality of high frequency closed-loop planning. Existing studies have shown that scene complexity and criticality in automated driving can be reasonably characterized by interpretable indicators such as traffic density, interaction intensity, topological complexity, and surrogate safety measures including TTC \cite{li2024complexity}. Meanwhile, event-triggered update strategies have been shown to reduce unnecessary computation in real time autonomous driving systems while maintaining comparable control performance \cite{zhou2023event}. Motivated by these observations, we replace the fixed asynchronous update strategy with a rule-based complexity-aware asynchronous scheduler.

Specifically, a lightweight scene complexity estimator is introduced on top of the scene encoder outputs and several readily available geometric and interaction statistics. At time step $t$, the scene complexity score is defined as
\begin{equation}
\begin{aligned}
C_t ={}& \alpha_1 \,\bar{N}_t
+\alpha_2 \,\bar{N}^{\mathrm{conf}}_t
+\alpha_3 \,\frac{1}{\overline{\mathrm{TTC}^{\min}_t}+\epsilon_{\mathrm{ttc}}} \\
&+\alpha_4 \,\mathbb{I}_{\mathrm{int}}(t)
+\alpha_5 \,\mathbb{I}_{\mathrm{nav}}(t)
+\alpha_6 \,\bar{\Delta}_t 
\end{aligned}
\end{equation}
where $\bar{N}_t$ denotes the normalized number of neighboring agents, $\bar{N}^{\mathrm{conf}}_t$ denotes the normalized number of potentially interactive or conflicting agents, \(\overline{\mathrm{TTC}^{\min}_t}\) denotes the normalized minimum TTC between the ego vehicle and surrounding agents, and \(\epsilon_{\mathrm{ttc}}\) is a small positive constant used to avoid numerical instability, $\mathbb{I}_{\mathrm{int}}(t)$ indicates whether the ego vehicle is currently located in a topologically complex area such as an intersection, merge, or turning region, $\mathbb{I}_{\mathrm{nav}}(t)$ indicates whether the current navigation command involves a high level maneuver change such as turning or lane changing, and $\bar{\Delta}_t$ measures the normalized short term scene variation, such as changes in surrounding agent states or traffic light conditions. All continuous complexity-related factors are normalized to comparable ranges before computing \(C_t\), and the coefficients $\alpha_i$ are set to equal values in the default configuration to assign the above factors the same prior importance. This setting keeps the scheduler interpretable and avoids introducing additional trainable parameters into the asynchronous update policy.

Based on the estimated complexity score, the semantic feature reuse length is adaptively adjusted according to a rule-based scheduling policy:
\begin{equation}
K_t^{\mathrm{sem}} =
\begin{cases}
K_{\min}, & C_t \ge \tau_h,\\
K_{\mathrm{mid}}, & \tau_l \le C_t < \tau_h,\\
K_{\max}, & C_t < \tau_l.
\end{cases}
\label{eq:semantic_reuse_step}
\end{equation}
where $\tau_h$ and $\tau_l$ denote the high and low complexity thresholds, respectively, and \(K_{\min}<K_{\mathrm{mid}}<K_{\max}\) denote the corresponding numbers of planning frames for reusing the most recent semantic feature. In our implementation, $\tau_l=0.35$ and $\tau_h=0.65$ are empirically selected according to the validation split statistics and the online latency budget, while \(K_{\min}=3\), \(K_{\mathrm{mid}}=9\), and \(K_{\max}=29\) frames are used to balance semantic freshness and online inference efficiency. Here, \(K_t^{\mathrm{sem}}\) is measured in frames, while \(\Delta t\) is reserved for the physical sampling interval used in trajectory state differentiation.
In this way, the LLM is invoked more frequently in highly interactive or rapidly changing scenes, while in simple or stable scenes the most recent semantic feature is reused for a longer period. Let the \(k\)-th semantic update time be \(t_k\), let the corresponding semantic feature be \(f_{\mathrm{sem}}^{(k)}\), and let \(K_k^{\mathrm{sem}}\) be the reuse length selected at this update time. The semantic feature used by the planning module can then be written as
\begin{equation}
\hat{f}_{\mathrm{sem}}^{\,t} = f_{\mathrm{sem}}^{(k)}, \qquad
t_k \le t < t_k + K_k^{\mathrm{sem}} 
\label{eq:semantic_feature_reuse}
\end{equation}

This complexity-aware scheduling strategy improves the trade-off between semantic richness and online efficiency. On the one hand, it avoids unnecessary LLM invocations in low complexity scenarios where high level semantics evolve slowly. On the other hand, it allows the semantic enhancement module to respond more promptly in high complexity scenarios involving dense interactions, topological changes, or increased collision risk. After the asynchronously updated semantic feature is obtained, it is injected into the downstream planning module and jointly used with scene encoding features for multimodal trajectory generation and surrounding agent behavior modeling. In this way, the planner can better capture high level semantic constraints and behavioral preferences associated with the current driving task, thereby improving planning performance in complex and long-tail traffic scenarios.

\subsection{IL-based Planning with Differentiable Optimization}

Within the proposed framework, we build a unified decision-making and planning module that combines IL-based planning with differentiable optimization. 
The IL planner generates initial ego trajectories and predicts surrounding agent behaviors from scene encoded features and asynchronously updated scene-associated instruction features. 
The differentiable optimizer then refines the selected ego trajectory using predicted agent trajectories and explicit residual costs, enabling joint learning of trajectory generation and constraint-aware refinement.

Following~\cite{cheng2024pluto}, the planning decoder represents multimodal driving behaviors using lateral and longitudinal queries. 
Reference-line polyline features are encoded as lateral queries, while longitudinal behaviors are modeled by learnable longitudinal query vectors. 
The fused queries are further processed by Query2Scene CrossAttn and LLM CrossAttn, where the former models road geometry, surrounding agents, and scene context, and the latter injects navigation intent and high level scene semantics from the asynchronous LLM feature.

To prevent the semantic branch from excessively interfering with the original planning backbone, we introduce an adaptive gate to control the semantic injection strength. 
Let the query at the \(l\)-th decoder layer be \(q^l\), the scene feature be \(F_{\text{scene}}\), and the adapted semantic feature be \(F_{\text{sem}}\). 
The query update is formulated as
\begin{equation}
\begin{aligned}
q^{l+1} ={}& \mathrm{CrossAttn}(q^l, F_{\text{scene}}, F_{\text{scene}}) \\
&+ g^l \cdot \mathrm{CrossAttn}(q^l, F_{\text{sem}}, F_{\text{sem}}),
\end{aligned}
\end{equation}
where \(g^l\) is the learnable gate at the \(l\)-th layer. 
The gate is initialized to zero, allowing the model to first rely on the original scene branch and then gradually learn the contribution of semantic guidance. 
The enhanced decoder features are fed into trajectory and score prediction heads to generate multimodal initial trajectories and confidence scores.

After initial trajectory generation and surrounding agent prediction, the selected ego trajectory is refined by a differentiable optimizer. 
The ego trajectory is denoted by \(u=\{u_t\}_{t=1}^{T_F}\), where \(u_t=\{x_t,y_t,\theta_t,v_t\}\). 
In our implementation, \(T_F=80\), and the optimizer operates on \(80\times4\) trajectory variables. 
The candidate trajectory predicted by the upstream IL planner is converted into the \((x,y,\theta,v)\) representation and used as the optimizer initialization.

Given the predicted surrounding agent trajectories \(\hat{s}\), reference-line information, and the current scene state, the final trajectory is obtained by solving
\begin{equation}
u^*=\arg\min_u \frac{1}{2}\sum_i \left\|\omega_i\, c_i(u,\hat{s})\right\|_2^2 ,
\end{equation}
where \(c_i(\cdot)\) is the \(i\)-th residual term and \(\omega_i\) is its weight. 
The residuals encode driving efficiency, comfort, safety, and kinematic feasibility as differentiable soft penalties. 
During the inner optimization, \(\hat{s}\) is fixed as the conditional prediction input and is updated at the next planning step.

The optimizer contains four categories of residual terms. 
First, for driving efficiency and reference-line consistency, the speed residual is
\begin{equation}
c_t^{\mathrm{speed}}=v_t-v_{\mathrm{limit},t},
\end{equation}
where \(v_{\mathrm{limit},t}\) is the local speed limit matched from the reference line. 
The position and heading residuals are
\begin{equation}
c_{\mathrm{lane\mbox{-}xy},t}=p_t-p_t^{\mathrm{ref}}, \quad
c_{\mathrm{lane\mbox{-}\theta},t}=\mathrm{wrap}\!\left(\theta_t-\theta_t^{\mathrm{ref}}\right),
\end{equation}
where \(p_t=(x_t,y_t)\), and \(p_t^{\mathrm{ref}}\) and \(\theta_t^{\mathrm{ref}}\) are the nearest matched reference-line position and heading. 
The heading residual is evaluated only at selected time steps to reduce optimization dimensionality and improve stability.

Second, for comfort, longitudinal acceleration and jerk are constrained. 
With \(a_t=(v_{t+1}-v_t)/\Delta t\) and \(j_t=(a_{t+1}-a_t)/\Delta t\), the residuals are
\begin{equation}
\begin{aligned}
c_{\mathrm{acc},t} &= 0.01|a_t|+\max(a_t-a_{\max},0)+\max(a_{\min}-a_t,0), \\
c_{\mathrm{jerk},t} &= 0.01|j_t|+\max(|j_t|-j_{\max},0),
\end{aligned}
\end{equation}
where \(a_{\max}=2.40~\mathrm{m/s}^2\), \(a_{\min}=-4.05~\mathrm{m/s}^2\), and \(j_{\max}=4.13~\mathrm{m/s}^3\), following the nuPlan criteria~\cite{caesar2021nuPlan}. 
The small \(L_1\) terms suppress unnecessary acceleration and jerk fluctuations.

Third, for safety, valid surrounding agents entering the local risk corridor of the ego trajectory are selected in the Frenet frame. 
The corridor is determined by longitudinal relation and lateral offset thresholds around the reference line, covering forward path-overlap conflicts and lateral or crossing agents entering the ego driving corridor. 
For each checked time step, the nearest or highest-risk target is used to define
\begin{equation}
c_{\mathrm{safety},t}=\max(d_t^{\mathrm{safe}}-d_t^{\mathrm{near}}, 0), \quad
d_t^{\mathrm{safe}} = \frac{L_{\mathrm{ego}} + L_{\mathrm{obj}}}{2} + 5.0 ,
\end{equation}
where \(d_t^{\mathrm{near}}\) is the Euclidean distance to the selected conflict target. 
The size term accounts for vehicle lengths, and the additional \(5.0\,\mathrm{m}\) provides a conservative urban safety buffer. 
Rear-end risks are mainly handled by the upstream interaction-aware prediction and planning module, closed-loop feedback, and safety-related evaluation metrics. 
This selective design balances collision-risk coverage and online efficiency.

Finally, for kinematic feasibility, planar consistency residuals are defined using a discrete bicycle model:
\begin{equation}
\begin{aligned}
c_{\mathrm{kin},t}^{x} &= (x_{t+1}-x_t)-v_t\cos\theta_t\,\Delta t, \\
c_{\mathrm{kin},t}^{y} &= (y_{t+1}-y_t)-v_t\sin\theta_t\,\Delta t .
\end{aligned}
\end{equation}
The heading rate and curvature are computed as \(\dot{\theta}_t=\mathrm{wrap}(\theta_{t+1}-\theta_t)/\Delta t\) and \(\kappa_t=\dot{\theta}_t/\max(v_t,1.0)\). 
The steering angle is \(\delta_t=\arctan(L\kappa_t)\), with \(L\) denoting the wheelbase. 
The steering and steering-rate residuals are \(c_{\mathrm{steer},t}=\mathrm{clip}(\delta_t)\) and \(c_{\Delta\mathrm{steer},t}=(\delta_{t+1}-\delta_t)/\Delta t\), which suppress infeasible steering actions and rapid steering variations.

The residual weights are generated by a lightweight cost weight network. 
The residual forms and physical thresholds are manually specified according to driving priors, while the network adjusts the relative importance of driving efficiency, comfort, safety, and kinematic feasibility. 
A fixed category-level scaling factor is also used to emphasize safety-related terms, preserving interpretability while allowing end-to-end weight adaptation.

For optimization, we employ the Levenberg--Marquardt solver in Theseus~\cite{pineda2022theseus}. 
The solver is initialized by the upstream predicted trajectory. 
At each iteration, residuals and Jacobians are computed through automatic differentiation, and trajectory variables are updated by solving the damped normal equation. 
We set the maximum number of LM iterations to 5, the step size to 0.2, and the absolute error tolerance to \(1\times10^{-3}\). 
The initial damping value is \(1\times10^{-1}\), and adaptive and ellipsoidal damping are enabled with a stability parameter of \(1\times10^{-6}\). 
During training, the executed LM iterations are retained in the computation graph, allowing gradients from the planning loss and optimization cost to propagate through residual evaluation, linear solving, and trajectory updates to the upstream planner.

\subsection{Learning Process}

We adopt a staged training strategy. 
For the LLM component, we follow the alignment assistance loss in~\cite{chen2024asyncdriver} for pretraining and fine tuning. 
After this stage, the LLM backbone is frozen during planner training, and the extracted scene-semantic features are used by the downstream planning module. 
The alignment assistance loss is formulated as
\begin{equation}
\begin{aligned}
\mathcal{L}_{\text{align}}
={}& \mathcal{L}_{1}(\tilde{x}_{va},x_{va})
+ CE(\tilde{x}_{dec},x_{dec})
+ CE(\tilde{x}_{traf},x_{traf}) \\
& + BCE(\tilde{x}_{adj},x_{adj})
+ BCE(\tilde{x}_{chg},x_{chg})
\end{aligned}
\end{equation}
where $\tilde{x}_{va}$, $\tilde{x}_{dec}$, $\tilde{x}_{traf}$, $\tilde{x}_{adj}$, and $\tilde{x}_{chg}$ denote the predictions of ego velocity and acceleration, future velocity decision, traffic light state, adjacent lane existence, and future lane change demand, respectively, and $x_{va}$, $x_{dec}$, $x_{traf}$, $x_{adj}$, and $x_{chg}$ are the corresponding labels.

For the planning module, the endpoint of the expert trajectory is projected onto the reference line to determine the target reference line and longitudinal query, yielding the target supervision trajectory $\hat{\tau}$ and its associated one-hot distribution $\pi_0^*$~\cite{cheng2024pluto}. 
The supervision trajectory is then passed through the differentiable optimizer to obtain the optimized trajectory $\tau^{*}$, which is supervised by the expert trajectory $\tau^{gt}$. 
The planning and confidence supervision terms are defined as
\begin{equation}
\mathcal{L}_{\text{plan}}
=
L_{1}^{\text{smooth}}(\tau^{*},\tau^{gt}),
\end{equation}
\begin{equation}
\mathcal{L}_{\text{cls}}
=
CE(\pi_0,\pi_0^*)
\end{equation}
where $\pi_0$ denotes the predicted confidence distribution over multimodal ego trajectory proposals, and $\pi_0^*$ denotes the target one-hot distribution. 
The IL supervision term is
\begin{equation}
\mathcal{L}_{\text{imi}}
=
\mathcal{L}_{\text{plan}}+\mathcal{L}_{\text{cls}} .
\end{equation}
Since $\tau^{*}$ remains differentiably connected to the initial trajectory proposal, the gradient of $\mathcal{L}_{\text{plan}}$ is propagated through the executed LM iterations to the upstream planning decoder and the cost weight generation network.

For surrounding agent prediction, we follow the prediction loss in PLUTO. 
Let $P_{1:N_A}$ denote the output of the prediction decoder and $P_{1:N_A}^{gt}$ denote the corresponding ground truth future trajectories. 
The prediction loss is
\begin{equation}
\mathcal{L}_{\text{pred}}
=
L_{1}^{\text{smooth}}(P_{1:N_A},P_{1:N_A}^{gt}) .
\end{equation}

Following~\cite{huang2023differentiable}, we also incorporate the overall optimization cost $\mathcal{L}_{\text{cost}}$, which is computed from the manually structured residual terms with weights produced by the cost weight generation network. 
The overall objective used in the staged training pipeline is
\begin{equation}
\mathcal{L}
=
\lambda_{\text{align}}\mathcal{L}_{\text{align}}
+
\lambda_{\text{pred}}\mathcal{L}_{\text{pred}}
+
\lambda_{\text{imi}}\mathcal{L}_{\text{imi}}
+
\lambda_{\text{cost}}\mathcal{L}_{\text{cost}}
\label{eq:overall_training_objective}
\end{equation}
where $\lambda_{\text{align}}$, $\lambda_{\text{pred}}$, $\lambda_{\text{imi}}$, and $\lambda_{\text{cost}}$ are the corresponding loss weights.

\section{EXPERIMENT}
\subsection{Experimental Setup}

To evaluate the proposed method from both benchmark and online system perspectives, experiments are conducted on the nuPlan closed-loop benchmark and a CARLA-ROS platform. The nuPlan benchmark provides reproducible quantitative comparison under the official closed-loop evaluation protocol, while the SIL platform validates online deployment and closed-loop execution under interactive simulation and cross module communication.

For standardized benchmarking, we evaluate the proposed method on the nuPlan dataset following its official closed-loop protocol. The closed-loop score considers collision avoidance, drivable area compliance, TTC margin, speed limit compliance, ride comfort, and route progress, providing a comprehensive assessment of planning performance in complex urban traffic environments~\cite{caesar2021nuPlan}. We compare the proposed method with representative baselines under the same evaluation setting, including IDM~\cite{treiber2000congested}, UrbanDriver~\cite{scheel2022urban}, GC-PGP~\cite{hallgarten2023prediction}, PDM-Closed~\cite{dauner2023parting}, PlanTF~\cite{cheng2024rethinking}, Diffusion Planner~\cite{zheng2025diffusion}, PLUTO~\cite{cheng2024pluto}, and AsyncDriver~\cite{chen2024asyncdriver}.

\begin{figure}[t]
    \centering
    \includegraphics[width=0.5\textwidth]{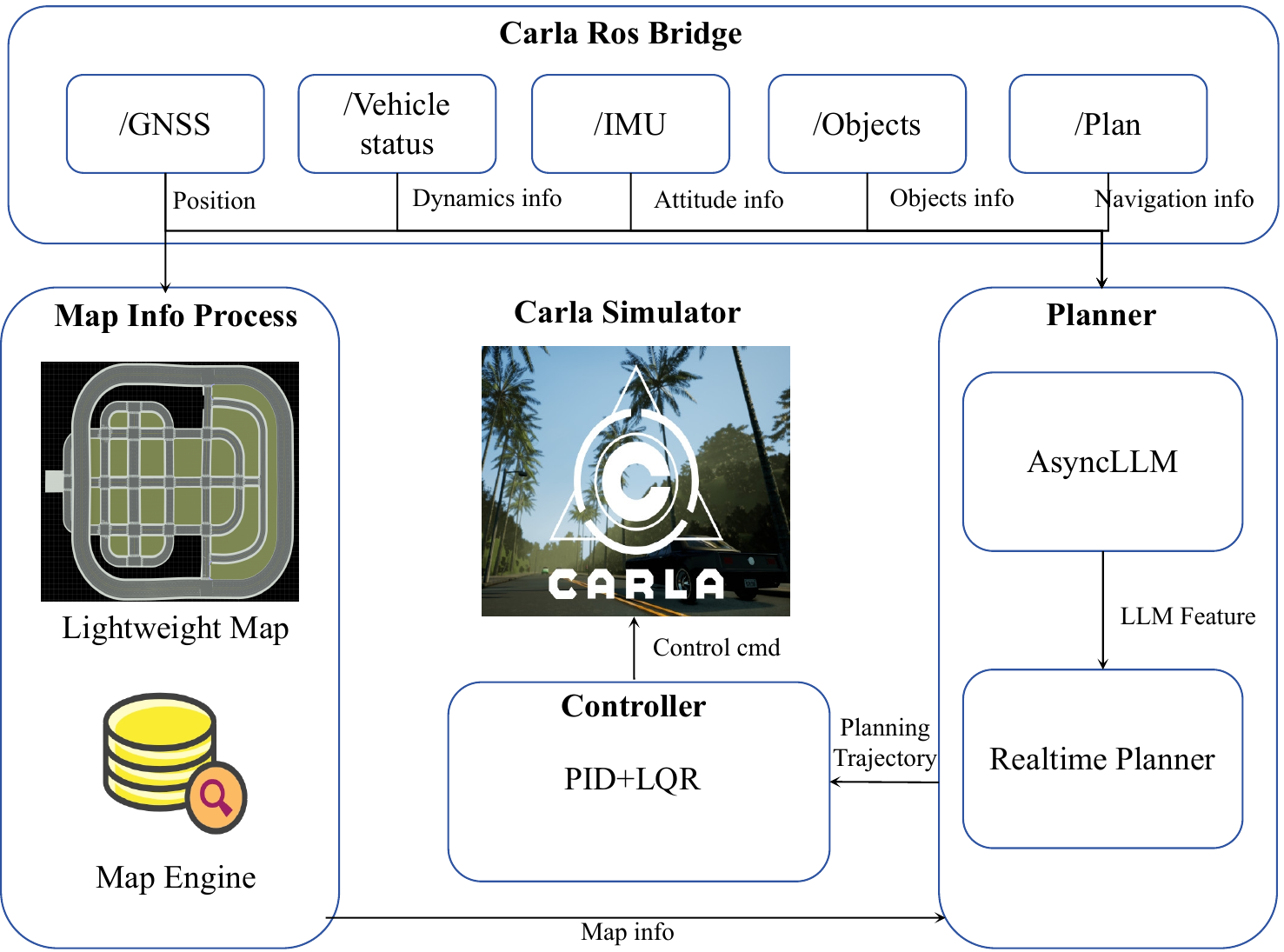}
    \caption{SIL platform communication architecture.}
    \label{fig:SIL}
\end{figure}

Beyond the benchmark evaluation, we conduct system level validation on a CARLA-ROS platform, whose communication architecture is shown in Fig.~\ref{fig:SIL}. CARLA provides the interactive traffic environment and publishes multi-source messages to ROS through Carla\_Ros\_Bridge, including IMU, GNSS, vehicle states, object states, and navigation information. The planner generates trajectories from these inputs, and the tracking controller converts them into control commands for CARLA execution. The LLM module and the real time planner are implemented as separate ROS nodes: the former publishes LLM features, while the latter performs trajectory planning inference upon receiving them. This decoupled design supports real time SIL validation.

\subsection{Implementation Details}

This subsection presents the main implementation details and key hyperparameter settings. The model is trained on an NVIDIA RTX 4090 GPU with a batch size of 128 for 25 epochs. The learning rate is linearly warmed up to \(1\times10^{-3}\) during the first 3 epochs and then decayed using a cosine schedule. For the LLM branch, we follow implementation in~\cite{chen2024asyncdriver} and adapt a LLaMA series language model with low rank adaptation (LoRA) using instruction style samples constructed from nuPlan training scenarios \cite{lin2026hsplitlora}. The original LLM backbone is frozen, while the LoRA parameters, feature adapter, adaptive gate, and downstream planning modules are trainable. The LLM branch runs in half precision during inference. In the loss function, the weights of all main task losses are set to 1.0, while the weight of the optimization-related auxiliary loss is set to \(1\times10^{-2}\). The remaining implementation hyperparameters are summarized in Table~\ref{tab:params}.

\begin{table}[t]
\centering
\caption{Key hyperparameter settings of the proposed method.}
\label{tab:params}
\small
\renewcommand{\arraystretch}{1.05}
\setlength{\tabcolsep}{4pt}
\begin{tabularx}{\columnwidth}{Xc}
\hline
\textbf{Parameter} & \textbf{Value} \\
\hline
Historical observation horizon $T_H$ & 20 \\
Future planning horizon $T_F$ & 80 \\
Number of surrounding traffic participants $N_A$ & 20 \\
Target vehicle screening radius (data augmentation) & 50 m \\
Dimension of instruction feature $F_{\text{sem}}$ & 128 \\
Low complexity threshold $\tau_l$ & 0.35 \\
High complexity threshold $\tau_h$ & 0.65 \\
Minimum semantic feature reuse length \(K_{\min}\) & 3 frames \\
Medium semantic feature reuse length \(K_{\mathrm{mid}}\) & 9 frames \\
Maximum semantic feature reuse length \(K_{\max}\) & 29 frames \\
\hline
\end{tabularx}
\end{table}

\section{RESULTS AND DISCUSSION}

\subsection{Comparison with Baselines}
Table~\ref{tab:baselines} reports the comparison results on the nuPlan closed-loop nonreactive challenge benchmark. The proposed method achieves the best overall score of 83.63, while also attaining the highest values on the safety-related metrics of Coll. and TTC, reaching 95.97 and 85.14, respectively. At the same time, it maintains strong performance on Drivable and Prog. These results indicate that the gain does not come from optimizing a single metric in isolation, but from a better overall balance among safety, trajectory feasibility, and driving efficiency. This can be attributed to three aspects. First, the surrounding agent centric data augmentation strategy increases the coverage of complex interactions and high value behavior samples, thereby improving the model's ability to learn from long-tail scenarios. Second, the differentiable optimization module jointly incorporates speed constraints, reference line consistency, safety margins, and kinematic constraints into trajectory refinement, and improves the alignment between trajectory generation and final execution objectives through end-to-end training. Third, the asynchronous LLM module further complements the planner with navigation intent and scene-semantic information, enabling the generated trajectories to better satisfy task requirements in complex scenarios.

The behavior of representative baselines further reflects the influence of different design choices. PLUTO performs strongly when equipped with post processing, whereas PLUTO w/o refine.\* drops substantially from 80.82 to 74.26, indicating that output stage refinement plays an important role in its closed-loop performance. AsyncDriver improves planning through asynchronous semantic enhancement, but its lack of explicit trajectory refinement limits its gains in safety-related and feasibility-related metrics. Diffusion Planner shows competitive trajectory generation ability, but it does not explicitly incorporate downstream residual-based optimization during planning. These comparisons suggest that combining high quality augmented supervision, semantic guidance, and differentiable trajectory refinement is beneficial for achieving more balanced closed-loop performance.

Table~\ref{tab:reac_baselines} presents the comparison results on the nuPlan closed-loop reactive challenges benchmark, where surrounding traffic participants can respond to the ego vehicle. The proposed method achieves the best overall score of 78.29 and obtains the best performance on Drivable, Direct., and TTC, with scores of 98.99, 99.58, and 88.44, respectively. Although PLUTO and PLUTO w/o refine.\* show slightly higher collision-related scores, the proposed method maintains competitive collision performance while achieving a better overall balance among drivable area compliance, driving direction consistency, TTC safety, and closed-loop progress. These results indicate that the proposed framework maintains robust behavior under dynamic interactions, benefiting from interaction-aware training samples, semantic guidance, and optimization-aware trajectory refinement.

\begin{table*}[t]
\centering
\caption{\textbf{Evaluation on nuPlan closed-loop nonreactive challenges on Hard20 split.} The best results are highlighted in \textbf{bold}, while the second best results are underlined with an \underline{underline} for clear distinction. \textit{Score}: average final score. \textit{Drivable}: drivable area compliance. \textit{Direct.}: driving direction compliance. \textit{Comf.}: ego is comfortable. \textit{Prog.}: ego progress along expert route. \textit{Coll.}: no ego at fault collisions. \textit{Lim.}: speed limit compliance. \textit{TTC}: time to collision within bound.}
\label{tab:baselines}

\setlength{\tabcolsep}{5pt}
\begin{tabular*}{0.92\textwidth}{@{\extracolsep{\fill}}lcccccccc@{}}
\toprule
\textbf{Method} & \textbf{Score} & \textbf{Drivable} & \textbf{Direct.} & \textbf{Comf.} & \textbf{Prog.} & \textbf{Coll.} & \textbf{Lim.} & \textbf{TTC} \\
\midrule
IDM                & 56.16             & 86.40             & 98.53             & 88.60             & 65.82             & 82.54             & 96.87             & 69.48             \\
GC-PGP              & 47.02             & 85.66             & 96.32             & 86.02             & 51.96             & 76.65             & 98.57             & 72.06             \\
UrbanDriver        & 51.67             & 88.60             & 95.96             & \textbf{97.79}    & 73.99             & 76.65             & 89.07             & 69.85             \\
PDM-closed         & 64.18             & 95.69             & \textbf{99.10}    & 77.06             & 68.20             & 87.81             & \textbf{99.57}    & 73.47             \\
PlanTF             & 72.56             & 94.48             & 96.69             & \underline{94.85} & 84.21             & 87.50             & 97.11             & 80.51             \\
PLUTO w/o refine.* & 74.26             & \textbf{97.76}    & 98.69             & 82.09             & 76.99             & 92.73 & \underline{98.61} & \underline{85.05} \\
AsyncDriver        & 74.27             & 92.47             & \underline{98.92} & 91.40             & 79.63             & 92.11             & 97.96             & 82.80             \\
Diffusion Planner  & 75.47             & 95.59             & 98.90             & 91.54             & 89.74             & 87.31             & 96.44             & 77.57             \\
PLUTO              & \underline{80.82} & 97.05             & 98.53             & 73.89             & \textbf{90.43}    & \underline{94.30} & 97.70             & 83.08             \\
\textbf{Ours}               & \textbf{83.63}    & \underline{97.43} & 98.90             & 84.24             & \underline{89.87} & \textbf{95.97}    & 97.38             & \textbf{85.14}    \\
\bottomrule
\end{tabular*}
\end{table*}

\begin{table*}[t]
\centering
\caption{\textbf{Evaluation on nuPlan closed-loop reactive challenges on Hard20 split.} The best results are highlighted in \textbf{bold}, while the second best results are underlined with an \underline{underline} for clear distinction.}
\label{tab:reac_baselines}

\setlength{\tabcolsep}{5pt} 
\begin{tabular*}{0.92\textwidth}{@{\extracolsep{\fill}}lcccccccc@{}}
\toprule
\textbf{Method} & \textbf{Score} & \textbf{Drivable} & \textbf{Direct.} & \textbf{Comf.} & \textbf{Prog.} & \textbf{Coll.} & \textbf{Lim.} & \textbf{TTC} \\
\midrule
IDM                & 62.26             & 84.19             & 98.53             & 87.87             & 69.60             & 84.38             & 96.52             & 72.43             \\
GC-PGP             & 44.29             & 85.29             & 96.88             & 88.60             & 46.51             & 83.82             & \textbf{98.63}    & 78.68             \\
UrbanDriver        & 49.06             & 82.72             & 95.59             & \textbf{98.52}    & \textbf{80.18}    & 69.85             & 86.14             & 63.97             \\
PDM-closed         & 62.26             & 84.19             & 98.53             & 87.87             & 69.60             & 84.38             & 96.52             & 72.43             \\
PLUTO w/o refine.* & 60.91             & 97.01             & 98.32             & 88.05             & 59.57             & \underline{93.47}             & 98.36             & 87.68             \\
PlanTF             & 60.34             & 94.49             & 97.43             & \underline{94.49} & 64.41             & 91.54             & 97.77             & 86.40 \\
AsyncDriver        & 64.89             & 94.62             & 98.82             & 81.36             & 67.31             & 85.48             & 98.09             & 73.48             \\
Diffusion Planner  & 68.50             & 95.22             & \underline{98.90} & 84.93             & \underline{76.72}             & 86.95             & 97.38             & 79.41             \\
PLUTO              & \underline{76.88} & \underline{97.39} & 98.32             & 89.55             & 74.67             & \textbf{94.59}    & \underline{98.58} & \underline{87.69} \\
\textbf{Ours}               & \textbf{78.29}    & \textbf{98.99}    & \textbf{99.58}    & 84.41             & 75.97             & 93.22             & 97.88             & \textbf{88.44}    \\
\bottomrule
\end{tabular*}
\end{table*}

\begin{figure*}[t]
    \centering
    \includegraphics[width=1.0\textwidth]{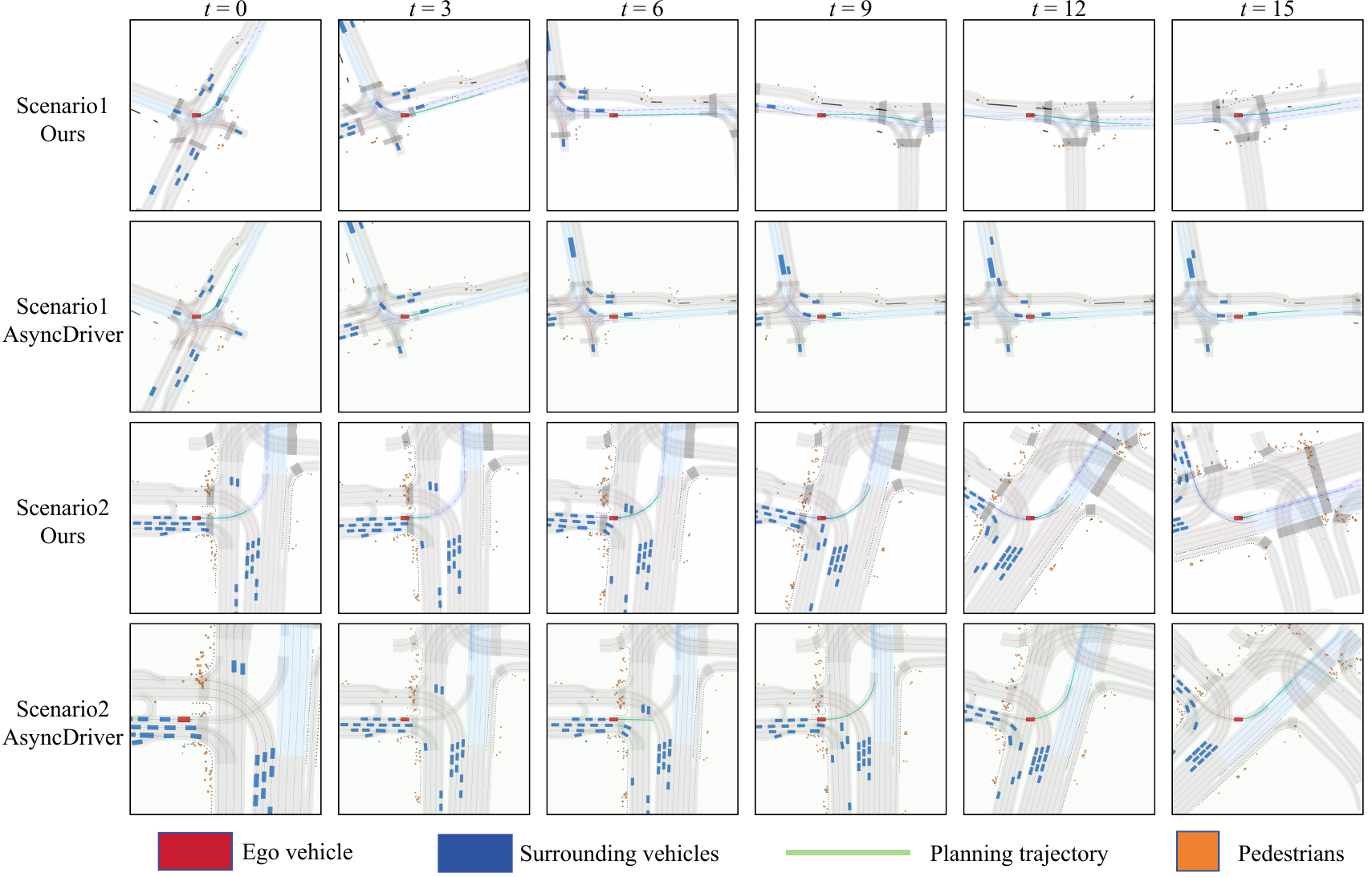}
    \caption{Qualitative comparison of closed-loop planning results on representative scenarios from the nuPlan Hard20 split. Snapshots are sampled from 0\,s to 15\,s. Compared with AsyncDriver, the proposed method generates more route consistent and drivable trajectories in complex intersection and turning scenarios, showing improved closed-loop robustness under dynamic interactions.}
    \label{fig:Qualitative_Results}
\end{figure*}

\begin{figure*}[t]
    \centering
    \includegraphics[width=1\textwidth]{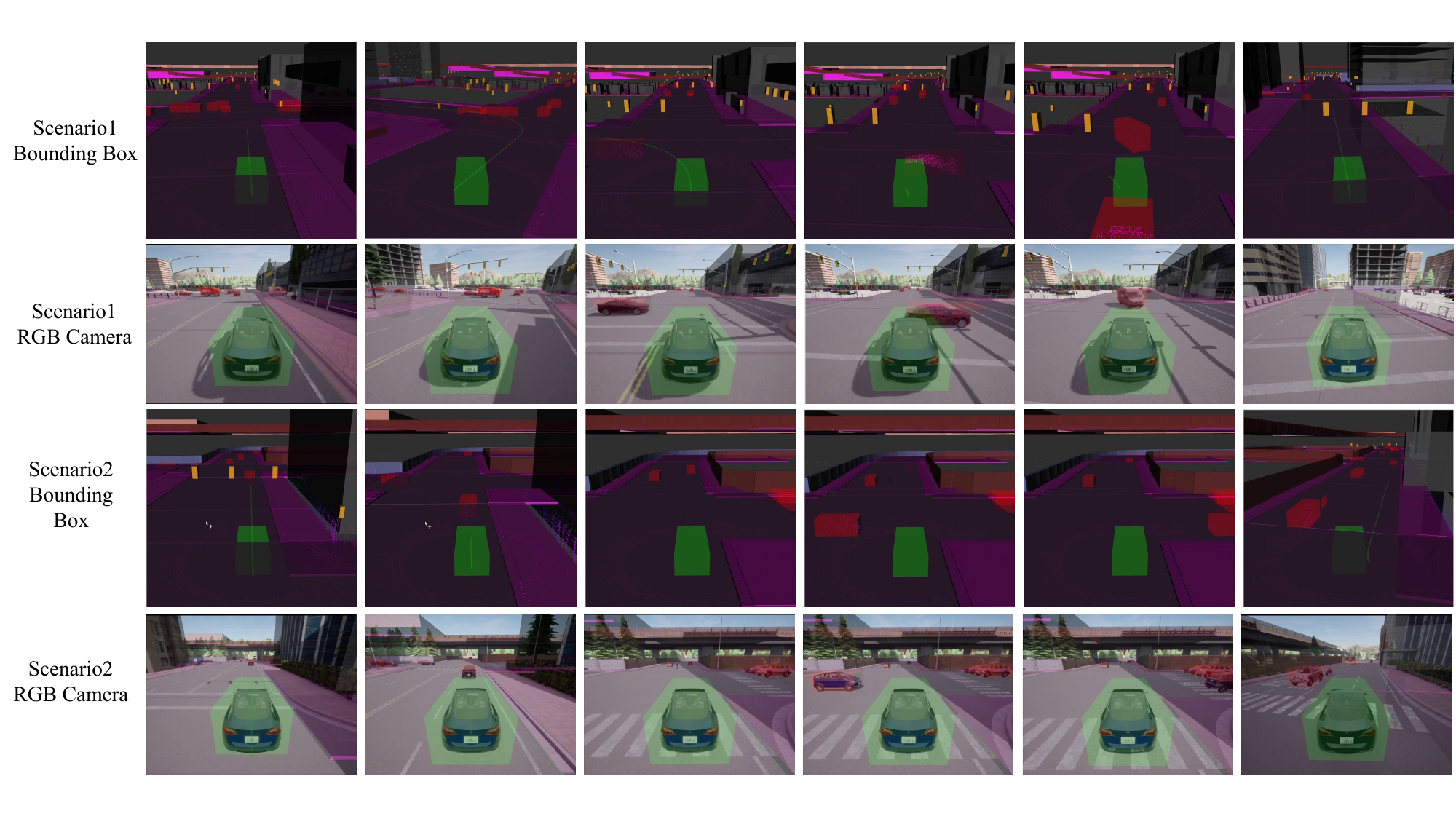}
    \caption{Qualitative closed-loop results on the CARLA-ROS platform, shown from both bounding box and RGB camera views. In representative urban scenarios, the proposed planner continuously generates executable trajectories under online feedback and maintains stable behavior when interacting with surrounding traffic participants, demonstrating its real time deployment feasibility in the SIL environment.}
    \label{fig:sil_pic}
\end{figure*}

\subsection{Qualitative Results}

To further analyze the behavior of the proposed method in complex interactive scenarios, Fig.~\ref{fig:Qualitative_Results} compares the proposed method with AsyncDriver in two representative left turn scenarios. The figure shows the closed-loop evolution from 0\,s to 15\,s, where the first and third rows correspond to the proposed method, and the second and fourth rows correspond to AsyncDriver.

In the first scenario, the target lane is occupied during the left turn. The proposed method adjusts its trajectory while preserving the left turn intention and completes a lane change during the turning process, thereby avoiding prolonged stopping and maintaining trajectory continuity. In contrast, AsyncDriver remains near the intersection for a longer period and fails to complete the maneuver adjustment in time. This difference indicates that the proposed method benefits not only from asynchronous semantic guidance, but also from differentiable trajectory refinement, which incorporates reference line consistency, safety margins, and kinematic feasibility to generate more executable adjustment trajectories.

In the second scenario, straight going vehicles continuously pass through the intersection during the left turn. AsyncDriver adopts a conservative waiting strategy and starts turning only after the intersection is nearly cleared. By contrast, the proposed method first moves moderately into the intersection, maintains the yielding relationship, and completes the turn once a suitable passing opportunity appears. This behavior shows that the proposed method can make dynamic rather than purely static yielding decisions. Such behavior benefits from surrounding agent centric data augmentation, high level semantic guidance, and constraint aware trajectory refinement, enabling higher passing efficiency while maintaining safety.

Beyond the nuPlan benchmark, the CARLA-ROS experiments further validate the online closed-loop execution capability of the proposed method, as shown in Fig.~\ref{fig:sil_pic}. Across the evaluated scenarios, the planner ROS node runs at 5\,Hz, achieving an 80\% success rate, a collision rate below 5\%, and an average traversal time of less than 30\,s. In the unprotected left turn scenario, the ego vehicle performs a preparatory lane change, yields to the oncoming straight going vehicle, and completes the turn after the conflict clears. In the right turn scenario, the ego vehicle handles right lane occupation and waits for the oncoming straight going vehicle before completing the maneuver. These results show that the proposed method can handle lane occupation, dynamic yielding, and turning maneuvers in continuous execution. This capability benefits from complexity-aware semantic enhancement and differentiable trajectory refinement, which jointly improve task awareness, safety, and trajectory executability on the SIL platform.

\begin{table*}[t]
\centering
\caption{Ablation study results on the effects of each component.}
\label{tab:ablation_component}
\small
\renewcommand{\arraystretch}{1.08}
\setlength{\tabcolsep}{4pt}
\begin{tabular}{lccccccccc}
\toprule
\textbf{Model} & \textbf{Description} & \textbf{Score} & \textbf{Drivable} & \textbf{Direct.} & \textbf{Comfort} & \textbf{Progress} & \textbf{Collisions} & \textbf{Speed} & \textbf{TTC}\\
\midrule
M0 & IL Planner 
& 60.43 & 96.85 & 98.23 & \textbf{88.19} & 58.87 & 93.50 & 98.48 & 88.58 \\

M1 & M0 + Data Aug. 
& 65.60 & \underline{97.58} & 98.39 & \underline{87.50} & 62.93 & \underline{94.15} & 98.26 & 88.71 \\

M2 & M1 + Diff. Opt. 
& 75.19 & 95.89 & 99.08 & 83.45 & 75.48 & \textbf{95.22} & \textbf{99.53} & 84.19 \\

M3$^{*}$ & M2 + AsyncLLM w/o adaptive gate 
& \underline{76.45} & 95.22 & \underline{99.52} & 84.69 & \underline{75.49} & 90.91 & 98.32 & 83.73 \\

M3 & M2 + AsyncLLM (Ours) 
& \textbf{78.29} & \textbf{98.99} & \textbf{99.58} & 84.41 & \textbf{75.97} & 93.22 & 97.88 & \textbf{88.44} \\
\bottomrule
\end{tabular}
\end{table*}

\begin{table}[t]
\centering
\caption{Ablation study results on different LLM update schedules.}
\label{tab:ablation_llm_schedule}
\small
\renewcommand{\arraystretch}{1.1}
\setlength{\tabcolsep}{4pt}
\begin{tabular}{lccc}
\toprule
\textbf{Model} & \textbf{Interval} & \textbf{Score} & \textbf{Runtime (ms) }\\
\midrule
Synchronous LLM            & every step & 80.46 & 477 \\
Fixed high frequency LLM   & 3 frames   & 79.61 & 237 \\
Fixed medium frequency LLM & 9 frames   & 76.74 & 155 \\
Fixed low frequency LLM    & 29 frames  & 75.33 & 126 \\
Complexity-aware AsyncLLM  & adaptive   & 78.29 & 172 \\
\bottomrule
\end{tabular}
\end{table}

\begin{table}[t]
\centering
\caption{Ablation analysis of the surrounding agent centric data augmentation strategy}
\label{tab:ablation_data_aug}
\small
\renewcommand{\arraystretch}{1.08}
\setlength{\tabcolsep}{5pt}
\begin{tabular}{llc}
\toprule
\textbf{Group} & \textbf{Setting} & \textbf{Score} \\
\midrule
Selection rule & Random selection & 61.35 \\
Selection rule & Interaction-aware selection & 65.60 \\
\midrule
Screening radius & \(r=15\,\mathrm{m}\) & 60.89 \\
Screening radius & \(r=25\,\mathrm{m}\) & 62.54 \\
Screening radius & \(r=50\,\mathrm{m}\) & 65.60 \\
Screening radius & \(r=75\,\mathrm{m}\) & 65.83 \\
Screening radius & \(r=100\,\mathrm{m}\) & 66.13 \\
\bottomrule
\end{tabular}
\end{table}

\begin{table}[t]
\centering
\caption{Ablation analysis of the complexity-aware scheduler.}
\label{tab:scheduler_sensitivity}
\small
\renewcommand{\arraystretch}{1.1}
\setlength{\tabcolsep}{4pt}
\begin{tabular}{lcc}
\toprule
\textbf{Setting} & \textbf{Score} & \textbf{Runtime (ms)} \\
\midrule
Default scheduler & 78.29 & 172 \\
Low thresholds \((0.25,0.55)\) & 78.83 & 224 \\
High thresholds \((0.45,0.75)\) & 75.63 & 132 \\
w/o TTC factor & 70.74 & 146 \\
w/o interaction factor & 72.57 & 138 \\
w/o navigation change factor & 75.21 & 155 \\
w/o scene variation factor & 76.46 & 163 \\
\bottomrule
\end{tabular}
\end{table}

\begin{table*}[t]
\centering
\caption{Ablation analysis of different residual categories in the differentiable optimizer.}
\label{tab:ablation_optimizer_residuals}
\small
\renewcommand{\arraystretch}{1.08}
\setlength{\tabcolsep}{4pt}
\begin{tabular*}{\textwidth}{@{\extracolsep{\fill}}cc*{8}{c}@{}}
\toprule
\textbf{Model} & \textbf{Removed residual category} & \textbf{Score} & \textbf{Drivable} & \textbf{Direct.} & \textbf{Comf.} & \textbf{Prog.} & \textbf{Coll.} & \textbf{Lim.} & \textbf{TTC} \\
\midrule
R0 & None, full residual set 
& \textbf{75.19} & 95.89 & \textbf{99.08} & \underline{83.45} & 75.48 & \textbf{95.22} & \textbf{99.53} & 84.19 \\

R1 & Efficiency residuals 
& 60.83 & \underline{96.85} & \underline{98.23} & \textbf{88.19} & 59.70 & \underline{93.90} & \underline{98.33} & \textbf{88.19} \\

R2 & Comfort residuals 
& \underline{66.93} & 94.35 & 97.18 & 50.01 & \underline{79.01} & 93.55 & 95.64 & \underline{85.48} \\

R3 & Safety residuals 
& 65.69 & \textbf{97.89} & 97.90 & 80.01 & \textbf{94.97} & 74.74 & 94.39 & 69.47 \\
\bottomrule
\end{tabular*}
\end{table*}

\subsection{Ablation Studies}

To analyze the contribution of key components and the sensitivity of important design choices, we conduct a series of ablation studies. All ablation studies in this subsection are conducted under the same nuPlan closed-loop reactive Hard20 setting, with only the target component or parameter changed in each study.

\subsubsection{Effects of Each Component} 
Table~\ref{tab:ablation_component} reports the stepwise ablation results of the key components. 
Adding the surrounding agent centric data augmentation strategy improves the score from 60.43 (M0) to 65.60 (M1), mainly with gains in Progress and Collisions, showing that reusing surrounding-agent trajectories helps learn interaction-related behaviors. 
Introducing the differentiable optimizer further increases the score to 75.19 (M2), with notable improvements in Progress, Collisions, and Speed. 
Although Comfort and TTC decrease slightly, the overall gain indicates that residual-based refinement improves closed-loop efficiency, safety, and executability. 
M3$^{*}$ adds the complexity-aware asynchronous LLM module without the adaptive gate and achieves 76.45, confirming the benefit of scene-associated semantic features. 
The full model M3 reaches 78.29 and outperforms M3$^{*}$ on Drivable, Direct, Progress, and TTC, suggesting that the adaptive gate reduces excessive semantic interference and enables controllable use of high-level guidance.

\subsubsection{Analysis of Data Augmentation Design}
Table~\ref{tab:ablation_data_aug} analyzes two key design choices in the surrounding agent centric data augmentation strategy. The default setting corresponds to M1 in Table~\ref{tab:ablation_component}. For target agent selection, interaction-aware selection improves the score from 61.35 to 65.60 compared with random selection, indicating that agents associated with lane changing, intersection traversal or turning, small TTC interactions, high lateral acceleration, and high magnitude speed behaviors provide more informative planning supervision. For the screening radius, increasing the radius from \(15\,\mathrm{m}\) to \(50\,\mathrm{m}\) improves the score from 60.89 to 65.60, showing that a small radius may miss useful interaction-relevant agents. Further enlarging the radius to \(75\,\mathrm{m}\) or \(100\,\mathrm{m}\) brings only marginal gains, with scores of 65.83 and 66.13, respectively. Therefore, we use \(50\,\mathrm{m}\) as the default radius to balance interaction coverage, sample relevance, and augmentation cost. These results show that the effectiveness of the proposed augmentation strategy mainly comes from selecting interaction-relevant surrounding agent trajectories rather than simply increasing the screening range.

\subsubsection{Effects of Different LLM Update Schedules}

Table~\ref{tab:ablation_llm_schedule} compares different LLM update schedules. Synchronous inference achieves the highest score of 80.46 but requires 477\,ms, while fixed-frequency updates reduce runtime at the cost of decreasing performance as the interval increases. The proposed complexity-aware AsyncLLM achieves a score of 78.29 with a runtime of 172\,ms, outperforming the fixed medium- and low-frequency variants with substantially lower overhead than synchronous and fixed high-frequency inference. These results demonstrate that adaptive semantic updates provide a practical balance between semantic freshness and online efficiency.

\subsubsection{Ablation Analysis of the Complexity-Aware Scheduler}
Table~\ref{tab:scheduler_sensitivity} shows that lowering the thresholds from \((0.35, 0.65)\) to \((0.25, 0.55)\) slightly improves the score from 78.29 to 78.83, but increases runtime from 172\,ms to 224\,ms. Conversely, higher thresholds \((0.45, 0.75)\) reduce runtime to 132\,ms but lower the score to 75.63. Removing any complexity factor also degrades performance, with the largest drops caused by excluding the TTC and interaction factors. These results confirm the importance of risk- and interaction-aware cues, while navigation-change and scene-variation cues provide complementary information. Overall, the default scheduler provides a practical balance between planning performance and online efficiency.

\subsubsection{Effects of Residual Categories}

Table~\ref{tab:ablation_optimizer_residuals} analyzes the three adjustable residual categories of efficiency, comfort, and safety, while the kinematic residuals are kept in all variants to maintain basic trajectory feasibility. 
The full residual set achieves the best overall score of 75.19. 
Removing the efficiency residuals (R1) reduces the score to 60.83 as Progress drops from 75.48 to 59.70, indicating overly conservative behavior. 
Removing the comfort residuals (R2) lowers Comfort to 50.01, confirming their role in maintaining motion smoothness. 
Without the safety residuals (R3), Progress increases to 94.97, whereas Collisions and TTC decrease to 74.74 and 69.47, respectively, revealing a substantial loss of interaction safety. 
These results demonstrate the complementary roles of the adjustable residual categories and the benefit of jointly balancing efficiency, comfort, and safety under basic kinematic feasibility constraints.

\section{Conclusions}

This paper presented a unified autonomous driving planning framework for IoT-enabled intelligent transportation scenarios by integrating surrounding agent centric data augmentation, asynchronous LLM-based semantic enhancement, and differentiable optimization into an IL-based planner. The proposed framework improves planning from three complementary aspects: interaction-oriented data augmentation, scene-aware semantic guidance, and optimization-aware trajectory refinement. Extensive experiments on the nuPlan closed-loop nonreactive and reactive benchmarks show that the proposed method achieves superior overall performance compared with representative baselines. Ablation studies and CARLA-ROS experiments further verify the effectiveness, efficiency, and online deployment potential of the proposed framework.

Future work will extend the framework in three directions. First, the current framework is mainly validated with structured intermediate representations and simulation-based closed-loop evaluation, while its robustness to real-world perception noise, localization drift, and system uncertainty requires further study. Second, although the asynchronous semantic module reduces LLM inference overhead through complexity-aware scheduling, the current rule-based scheduler may have limited adaptability in more open and diverse traffic environments. Third, the differentiable optimization module mainly uses soft residual penalties for safety, comfort, efficiency, and basic vehicle kinematics. In particular, the safety residual adopts a selective local risk formulation for online efficiency, and the kinematic residuals use a lightweight bicycle model formulation for trajectory refinement. Future work will investigate richer multi-directional and multi-agent interaction modeling, especially for rear-end and dense interaction cases, as well as dynamic vehicle constraints for aggressive or high-speed maneuvers. We will also explore more adaptive semantic scheduling, multimodal semantics and V2X integration, and higher fidelity hardware-in-the-loop and real vehicle validation.

\bibliographystyle{IEEEtran}
\normalem
\bibliography{IEEEabrv,references}

\end{document}